\documentclass[runningheads]{llncs}
\usepackage{graphicx}
\usepackage{xcolor}
\usepackage{comment}
\usepackage[utf8]{inputenc}
\usepackage{multirow}
\usepackage{arydshln}
\usepackage{subfig}
\usepackage{float}
\usepackage{array}
\usepackage{tabularx}
\usepackage{graphicx}
\usepackage{url}
\usepackage{adjustbox}
\usepackage{xspace}
\usepackage{xcolor}
\usepackage{comment}
\usepackage{caption} 
\usepackage{booktabs}
\PassOptionsToPackage{bookmarks=false}{hyperref}
\usepackage{nicematrix}
\restylefloat{table}
\usepackage{latexsym}
\usepackage{amsmath, amssymb}
\usepackage{type1cm}        
\usepackage{makeidx}         
\usepackage{graphicx}        
\usepackage{multicol}        
\usepackage[bottom]{footmisc}

\usepackage{newtxtext}       %
\usepackage[varvw]{newtxmath} 
\usepackage{type1cm}        
\usepackage{makeidx}         
\usepackage{graphicx}        
\usepackage{multicol}        
\usepackage[bottom]{footmisc}

\usepackage{newtxtext}       %
\usepackage[varvw]{newtxmath}       

\makeindex 

\begin{document}
\title{An Analysis of Abstractive Text Summarization Using Pre-trained Models}

%
%
\author{Tohida Rehman\inst{1}\thanks{corresponding author} \and Suchandan Das\inst{1} \and
Debarshi Kumar Sanyal \inst{2}\and\\ Samiran Chattopadhyay\inst{3}}

\authorrunning{Rehman, et al.}
\titlerunning{Rehman, et al.}
%

\institute{Jadavpur University, Kolkata, India.\\ \email{\{tohida.rehman, suithit\}@gmail.com}
\and
Indian Association for the Cultivation of Science, Kolkata, India.\\
\email{debarshisanyal@gmail.com} \and Institute for Advancing Intelligence (IAI), TCG CREST, Kolkata, India;\\ Jadavpur University, Kolkata, India.\\
\email{samirancju@gmail.com}}

\maketitle

\begin{abstract}\unskip
People nowadays use search engines like Google, Yahoo, and Bing to find information on the Internet. Due to explosion in data, it is helpful for users if they are provided relevant summaries of the search results rather than just links to webpages. 
Text summarization has become a vital approach to help consumers swiftly grasp vast amounts of information. 
In this paper, different pre-trained models for text summarization are evaluated on different datasets. Specifically, we have used three different pre-trained models, namely, google/pegasus-cnn-dailymail, T5-base, facebook/bart-large-cnn. We have considered three different datasets, namely, CNN-dailymail, SAMSum and BillSum to get the output from the above three models. The pre-trained models are compared over these different datasets,  each of 2000 examples, through ROUGH and BLEU metrics.


\end{abstract}

\section{Introduction}
Given a long text, humans have a natural tendency to remember its most important points in a summary form.
The volume of data around us is growing to the point that we need to find a solution that will deliver accurate and timely summary information. It requires a tool or approach for extracting an accurate summary from a large amount of data. Automatic text summarization is an example of a technology that may be used to achieve this. Broadly, there are two approaches for summarization \cite{luhn1958automatic,radev2002introduction}: \textit{extractive} and \textit{abstractive}. \textit{Extractive methods} \cite{knight2002summarization,jing2000cut,radev2002introduction} generally copy whole sentences from the input source text and combine them into a summary, throwing away the unimportant sentences from the input. \textit{Abstractive methods} \cite{genest2012fully} can generate novel words during summarization similar to how a human being does the job, that is, first reads the whole document, understands it, then summarizes by inducing new and appropriate words.
The main contributions of this paper are: (1) Evaluating the performance of three different pre-trained models, namely, google/pegasus-cnn-dailymail, T5-base, facebook/bart-large-cnn on CNN-dailymail \cite{hermann2015teaching,nallapati2016abstractive}, SAMSum \cite{gliwa2019samsum}, BillSum \cite{kornilova2019billsum} datasets. We record significant results using the three pre-trained models. Performance evaluation of the pre-trained models has been done with ROUGH \cite{lin2004rouge} and BLEU \cite{papineni2002bleu} metrics.   

\section{Literature Review} For abstractive summarization, Nallapati et al. \cite{nallapati2016abstractive} has proposed an a model based an Attentional Encoder Decoder Recurrent Neural Networks for an abstractive text summarization. Performance of basic encoder and decoder model has been improved through Bahdanau et al. \cite{bahdanau2014neural}. Luong et al. \cite{luong2015effective} proposed attention mechanism:
a global approach which always attends to all source words, and a local one that only looks at a subset of source words at a time. See et al. \cite{see2017get} offer a detailed study of numerous abstractive text summarization models using pointer-generator networks. Sutskever et al. \cite{sutskever2014sequence} offer a multilayer LSTM based end-to-end solution to sequence learning; here, the input for the encoder is a fixed length of text. Lin et al. \cite{lin2018global} has proposed a global encoding mechanism of abstractive text summarization. This paper uses GRU based encoder and decoder with one extra attention layer. Shi et al. \cite{shi2021neural} has proposed a seq2seq model to improve performance. This has improved fluency and human readability, generated high-quality summaries and captured the salient information.
{Aksenov et al. \cite{aksenov2020abstractive} has proposed  a new method called BERT windowing.} This method helps to process a long text (whose length is longer than BERT window) in a chunk-wise manner. 
Most of these techniques differ in one of these three categories: network structure, parameter inference, and decoding/generation nature.  

\section{Datasets}
Text summarization can be applied to a variety of datasets. Three different datasets are used in this paper: CNN-dailymail \cite{hermann2015teaching,nallapati2016abstractive}, SAMSum \cite{gliwa2019samsum} and BillSum \cite{kornilova2019billsum}. 

\textbf{CNN-dailymail} dataset is an English-language dataset with little over 300,000 unique news items published by CNN and Daily Mail writers. Version 3.0.0, which can be used to train both abstractive and extractive summarization, was employed in this study.
News stories and highlight sentences make up the dataset pair. The articles are utilised as the context in the question-answering setting, and entities are concealed one at a time in the highlight sentences, resulting in cloze-style problems in which the model correctly identifies which entity in the context has been hidden in the highlight. The highlight sentences are concatenated in the summarising setting to produce an article summary. 
The model generated summary's are compared to the written highlights to find out the ROUGE \cite{lin2004rouge}  score and BLEU \cite{papineni2002bleu} score for a specific article. 

\textbf{SAMSum} dataset consists about 16k messenger-like chats with summaries.
Linguists conversant in English created and recorded the conversations. The statistic represents the percentage of real-life issues. It has 16369 talks that are evenly split between four groups.
Data field consists of text of dialogue or conversation strings, summary form of the dialogue or conversations written by human and unique id of each example. 
In split, there are 14,732 instances in the train set, 818 in the validation set, and 819 in the test set. The summaries are annotated, with the expectation that they will be 1) brief, 2) extract key bits of information, 3) include interlocutors' names, and 4) written in the third person. There is only one summary reference in each dialogue or conversation. Samsung R\&D Institute, Poland prepared and published this dataset for non-commercial research purposes.

\textbf{BillSum} dataset summaries state bills in the United States Congress and California.
It has the following attributes: 1) text: the text of the bill 2) synopsis/summary: a synopsis/summary of the legislation, 3) title: the bills titles, 4) text  len is the number of character in the text, and 5) sum len is the number of characters in the summary. A text string,  summary string, and a title string make up the data field. Legislation bills have complex sentence structure and nature. Hence, it is a challenging, important and useful dataset for text summarization.

For comparison of the results of the three different pre-trained models, we have selected a set of 2000 test examples from each of the above datasets.

\section{Pre-trained models}
Zhang et al. \cite{zhang2020pegasus} introduced the Pegasus model in \textbf{PEGASUS}: ``\textit{Pre-training with Extracted Gap-sentences for Abstractive Summarization''}.
Pegasus pre-training task is designed to be comparable to summarization: key sentences from an input document are removed/masked, and the remaining phrases are combined into one output sequence, similar to an extractive summary.
Pegasus achieves good summary performance as evaluated by ROUGE and BLEU metrics in all 12 downstream datasets.

Raffel et al. \cite{raffel2019exploring} introduced the \textbf{ T5-base} model in \textit{``Exploring the Limits of Transfer Learning with a Unified Text-to-Text Transformer''}. \textbf{T5-base} model is an encoder-decoder model that has been pre-trained on a multi-task combination of unsupervised and supervised workloads, with each task transformed to text-to-text. T5-base works well on a range of jobs right out of the box by prefixing each task's input with a distinct prefix. T5 uses relative scalar embeddings. Encoder input padding can be done on the left and on the right. T5 comes in different sizes, here we used only t5-base only.

Lewis et al. \cite{lewis2019bart} introduced the \textbf{BART} (Bidirectional and Auto-Regressive
Transformers) \textbf{model}: \textit{``Denoising Sequence-to-Sequence Pre-training for Natural Language Generation, Translation and Comprehension''}. BART is a transformer encoder-encoder (seq2seq) model with a bidirectional (BERT( Bidirectional Encoder Representations from Transformers)-like) encoder and an autoregressive (GPT(Generative Pre-trained Transformer)-like) decoder.
It is implemented as a sequence-to-sequence model with a bidirectional encoder over corrupted text and a left-to-right autoregressive decoder\cite{lewis2019bart}. 
The pre-training job entails rearranging the sequence of the original phrases at random and using a new in-filling method that replaces text spans with a single mask token.
When fine-tuned for text production, BART is especially successful in  summarization as well as comprehension tasks.

\subsection{Pre-trained model comparison }
In this sub-section, we compare several pre-trained summarizers.
We choose google/pegasus-cnn-dailymail, T5-base, and facebook/bart-large-cnn models after going through the abstractive text summarization models on the hugging face website\footnote[1]{\url{https://huggingface.co/google/pegasus-cnn_dailymail}} \footnote[2]{\url{https://huggingface.co/t5-base}} \footnote[3]{\url{https://huggingface.co/facebook/bart-large-cnn}}.
Rather than looking for models that were trained on the same set of data, we selected models that were trained on different text corpora. We are interested to analyze how these models perform on a common set of test documents.

To compare these models effectively, they were all initialized and put into a Python dictionary for subsequent use.
The models were given the same articles from each of the datasets. 
The pre-trained models have character restrictions for that each and every model has some predefined number of accepted tokens in a sequence like 512 tokens only. When attempting to produce summaries for each article in the datasets, errors are generated if the restrictions are not obeyed. To address it, we have used the concept of BERT windowing \cite{aksenov2020abstractive}.
Some BERT based pre-trained models in the literature condense a lengthy text by eliminating everything except what it considers to be the most important sentences. We followed the same approach to reduce the input size; after an article has been reduced, an abstractive summary is created using the pre-trained models.

\section{Evaluation}
After the pre-trained models have summarised all the articles, the summaries are evaluated using ROUGE and BLEU metrics. 
When comparing the produced summaries to the reference summaries used for assessment, ROUGE determines the precision, recall, and f-measure for each model. Here, TP-True positive, FP-False positive, FN-False Negative naming conventions used.\\ 
Recall (r) is defined as:
\begin{equation}
    r = TP/(TP+FN)
\end{equation}
Precision (p) is defined as:
\begin{equation}
    p = TP/(TP+FP)
\end{equation}
F-measure (f) values are calculated using the formula:
\begin{equation}
    f = 2*((r*p)/(r+p))
\end{equation}
We have used another metric called BLEU to evaluate the performance of the models. BLEU (bilingual assessment understudy) is a method to assess the type of text produced by a model. BLEU is a metric that for ensuring a link to human quality choices, and it is still one of the most well-known automated and minimum cost estimates today. It gives a value between 0 and 1. It represents how much close the predicted summary is to the reference summary.

Using these measures, we are able to assess how well each model summarised the data. 
Despite the lack of GPU support in our computers, we have been able to test the three pre-trained models using 3 different dataset. We have successfully compared and assessed all the 3 models.

\begin{figure*}[!htb] 
\centering 
\includegraphics[width=12.2cm,height=7.6cm]{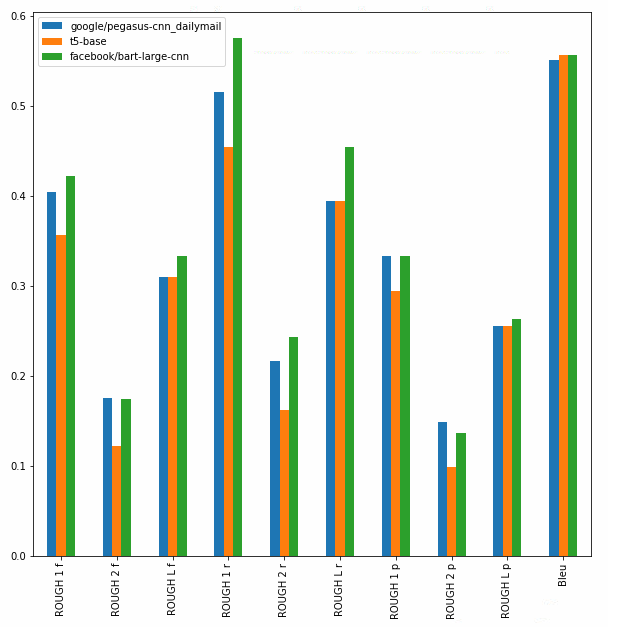}
\caption{Performance of three pre-trained models using CNN-dailymail dataset.}
\centering
\label{fig:con2fig1}
\end{figure*}
\begin{figure*}[!htbp] 
\centering 
\includegraphics[width=12.2cm,height=7.6cm]{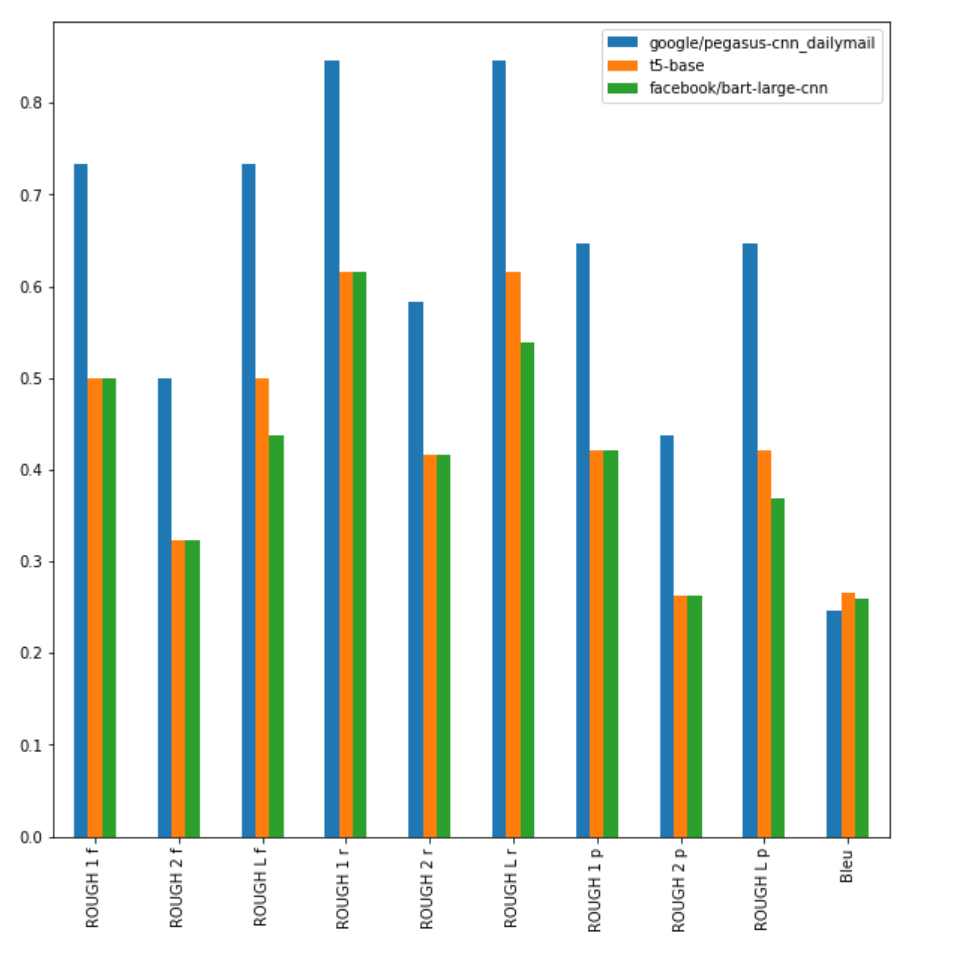}
\caption{Performance of three pre-trained models using SAMSum dataset.}
\centering
\label{fig:con2fig2}
\end{figure*}
\begin{figure}[!htb] 
\centering 
\includegraphics[width=12.2cm,height=7.6cm]{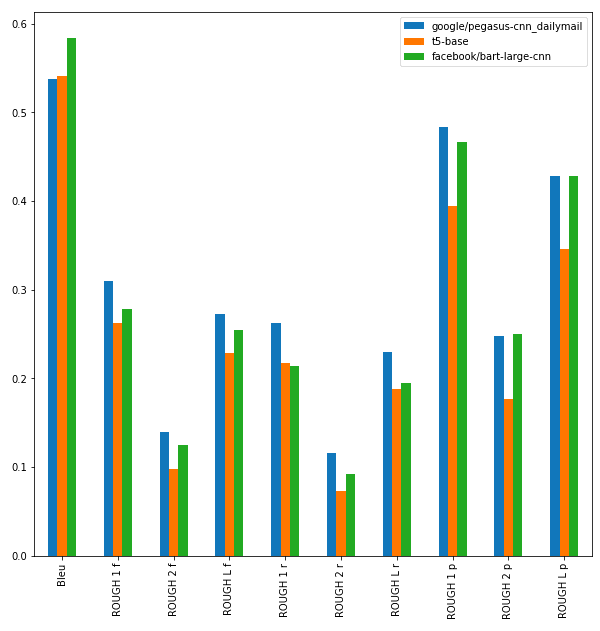}
\caption{Performance of three pre-trained models using BillSum dataset.}
\centering
\label{fig:con2fig3}
\end{figure}

The observation for \textbf{CNN-dailymail} dataset are shown in Table \ref{Table:table1}. Figure \ref{fig:con2fig1} shows the graphical representation of their outputs. The observation for \textbf{SAMSum} dataset are shown in Table \ref{Table:table2}. Figure \ref{fig:con2fig2} shows the graphical representation of their outputs. The observation for \textbf{BillSum} dataset are shown in Table \ref{Table:table3}. Figure \ref{fig:con2fig3} shows the graphical representation of their outputs. (Note that in Figures \ref{fig:con2fig1}, \ref{fig:con2fig2} and \ref{fig:con2fig3} where the X-axis contains the labels ROUGE\textit{1f} (f-measure for ROUGE-1), ROUGE\textit{2f}, ROUGH\textit{Lf}, ROUGH\textit{1r} (recall for ROUGE-1), ROUGH\textit{2r}, ROUGH\textit{Lr}, ROUGE\textit{1p} (precision for ROUGE-1), ROUGE\textit{2p},  ROUGE\textit{Lp}, and BLEU, and the color codes of the vertical bars are as follows: blue: google/pegasus-cnn-dailymail, orange: T5-base, green: facebook/bart-large-cnn.)

\begin{table*}[!htb]
\centering
\begin{tabular}{|c|c|c|c|c|c|c|c|c|c|c|}\hline
\multirow{2}{*}{}&\multicolumn{9}{|c|}{ROUGE} &\multirow{3}{*}{BLEU}\\\cline{2-10}
&\multicolumn{3}{|c|}{ROUGE-1} & \multicolumn{3}{|c|}{ROUGE-2}&\multicolumn{3}{|c|}{ROUGE-L} & \\\hline
&R&P&F1&R&P&F1&R&P&F1&\\\hline
\textbf{google/pegasus-cnn-dailymail} &40.22 &32.15 &34.60 &17.64 &13.98 &14.97 &37.51 &30.06 &32.32 &55.09\\\hline
\textbf{T5-base} &33.96 &30.35 &31.10 &13.37 &11.67 &12.00 &31.43 &28.10 &28.79 &55.70\\\hline
\bf{facebook/bart-large-cnn} &\bf{41.76} &\bf{33.69} &\bf{36.45} &\bf{18.67} &\bf{14.60} &\bf{15.92} &\bf{38.84} &\bf{31.36} &\bf{33.92} &\bf{55.71} \\\hline
\end{tabular}
\vspace*{1mm}
\caption{Observations of ROUGE and BLEU performance using CNN-dailymail dataset.}
\label{Table:table1}
\end{table*}
\vspace{-4em}
\begin{table*}[!htb]
\centering
\begin{tabular}{|c|c|c|c|c|c|c|c|c|c|p{1cm}|} \hline
\multirow{2}{*}{}&\multicolumn{9}{|c|}{ROUGE} &\multirow{3}{*}{BLEU}\\\cline{2-10}
&\multicolumn{3}{|c|}{ROUGE-1}&\multicolumn{3}{|c|}{ROUGE-2}&\multicolumn{3}{|c|}{ROUGE-L}&\\\hline
&R&P&F1&R&P&F1&R&P&F1&\\\hline
\textbf{google/pegasus-cnn-dailymail} &38.26 &25.12 &28.39 &10.69 &7.24 &8.05 &33.76 &22.63 &25.36 &24.59\\\hline
\textbf{T5-base} &33.11 &26.44 &27.88 &9.16 &7.66 &7.72 &29.38 &23.60 &24.82 &\bf{26.54}\\\hline
\textbf{facebook/bart-large-cnn} &\bf{40.63} &\bf{30.57} &\bf{33.02} &\bf{13.16} &\bf{10.10} &\bf{10.58} &\bf{36.38} &\bf{27.69} &\bf{29.78} &25.95\\\hline
\end{tabular}
\vspace*{1mm}
\caption{Observations of ROUGE and BLEU performance using SAMSum dataset.}
\label{Table:table2}
\end{table*}
\vspace{-4em}
\begin{table*}[!htb]
\centering
\begin{tabular}{|c|c|c|c|c|c|c|c|c|c|p{1cm}|}\hline
\multirow{2}{*}{}&\multicolumn{9}{|c|}{ROUGE} &\multirow{3}{*}{BLEU}\\\cline{2-10}
&\multicolumn{3}{|c|}{ROUGE-1}&\multicolumn{3}{|c|}{ROUGE-2}&\multicolumn{3}{|c|}{ROUGE-L}&\\\hline
&R&P&F1&R&P&F1&R&P&F1&\\\hline
\textbf{google/pegasus-cnn-dailymail} &\bf{26.30} &\bf{48.41} &\bf{30.95} &\bf{11.59} &24.82 &\bf{13.99} &\bf{22.95} &42.78 &\bf{27.25} &\bf{53.81}\\\hline
\textbf{T5-base} &21.71 &39.45 &26.29 &7.31 &17.71 &9.73 &18.77 &34.56 &22.91 &54.11\\\hline
\textbf{facebook/bart-large-cnn} &21.43 &46.65 &27.82 &9.20 &\bf{24.99} &12.52 &19.53 &\bf{42.84} &25.47 &58.41\\\hline
\end{tabular}
\vspace*{1mm}
\caption{Observations of ROUGE and BLEU performance using BillSum dataset.}
\label{Table:table3}
\end{table*}
\vspace*{-5mm}

\subsection{Sample Output} 
Figure \ref{fig:output1} illustrates the effectiveness of the 3 pre-trained models on  CNN-dailymail dataset.  Figure \ref{fig:output2} illustrates the effectiveness of the 3 pre-trained models on  SAMSum dataset. Figure \ref{fig:output3} illustrates the effectiveness of the 3 pre-trained model on  BillSum dataset. The google/pegasus-cnn-dailymail model and the facebook/bart-large-cnn model generate better summary on  CNN-dailymail dataset and SAMSum dataset while  the google/pegasus-cnn-dailymail model generates better summary on  BillSum dataset.

\begin{figure*}[tbp] 
\centering 
\begin{tabular}{ |p{12.2cm}|} \hline
\textbf{Article(truncated):} CNN Emergency operators get lots of crazy calls but few start like this. Caller Hello I'm trapped in this plane and I called my job but I'm in this plane. Operator You're where? Caller I'm inside a plane and I feel like it's up moving in the air. Flight 448 can you please tell somebody to stop it. The frantic 911 call came just as the Alaska Airlines flight had taken off from Seattle Tacoma International Airport on Monday afternoon. The caller was a ramp agent who fell asleep in the plane's cargo hold. The cell phone call soon broke up but the man was making himself known in other ways as the crew and passengers reported unusual banging from the belly of the Boeing 737. The pilot radioed air traffic control and said he would make an emergency landing. There could be a person in there so we're going to come back around he told air traffic control. The ramp agent who took the untimely nap and caused all the fuss is an employee of Menzies Aviation a contractor for Alaska Airlines that handles loading the luggage. He'll no longer have the option of dozing aboard one of the airline's planes. The Menzies employee has been permanently banned from working on Alaska Airlines planes said Bobbie Egan a spokeswoman for the airline. Flight 448 which was on its way to Los Angeles only spent 14 minutes in the air. Other than being scared the agent never was in any real danger. The cargo hold is pressurized and temperature controlled the airline said. The passengers knew something wasn't right almost as soon as the plane took off. All of a sudden we heard all this pounding underneath the plane and we thought there was something wrong with the landing gear Robert Higgins told CNN affiliate KABC. Not everyone heard the banging but it was soon clear this wasn't a normal flight. We just took off for L.A. regular and then ... about five minutes into the flight the captain came on and said we were going back and we'd land within five to seven minutes and we did passenger Marty Collins told affiliate KOMO. When we landed was when all the trucks and the police and the fire trucks surrounded the plane. I think it’s scary and really unsafe too Chelsie Nieto told affiliate KCPQ. Because what
if it’s someone who could have been a terrorist? The employee started work at 5 a.m. and his shift
was scheduled to end at 2 30 p.m. just before the flight departed. The agent was off the two days
prior to the incident and had taken a lunch break and a break in the afternoon before making his
way into the cargo hold according to a source familiar with the investigation. The man had been
on a four person team loading baggage onto the flight. During a pre departure huddle the team
lead noticed the employee was missing. The team lead called into the cargo hold for the employee
and called and texted the employee’s cell phone but did not receive an answer. His co workers
believed he finished his shift and went home the airline’s blog said. It’s believed he was hidden by
luggage making it difficult for the rest of his team to see him the source said. All ramp employees
have security badges and undergo full criminal background checks before being hired according
to the airline. After the delay the flight with 170 passengers and six crew members on board made
it to Los Angeles a couple of hours late. CNN’s Dave Alsup Joshua Gaynor and Greg Morrison
contributed to this report.\\\hline 	    
{\bf Human-written summary:} The ramp agent fell asleep in the plane's cargo hold . He can no longer work on Alaska Airlines flights . \\\hline 	    
{\bf Predicted summary with google/pegasus-cnn-dailymail:} The frantic 911 call came just as the Alaska Airlines flight had taken off .<n>The caller was a ramp agent who fell asleep in the plane's cargo hold .<n>The pilot radioed air traffic control and said he would make an emergency landing .\\\hline 	    
{\bf Predicted summary with T5-base:} a ramp agent fell asleep in the plane's cargo hold . the caller is an employee of a contractor for Alaska airlines . he'll no longer have the option of dozing aboard one of the airline's planes .\\\hline 	   
{\bf Predicted summary with facebook/bart-large-cnn:} A ramp agent fell asleep in the cargo hold of an Alaska Airlines flight. The crew and passengers reported unusual banging from the belly of the Boeing 737. The pilot radioed air traffic control and said he would make an emergency landing. \\\hline
\end{tabular} 	
\caption{Original news article, human-written summary, and the summaries produced by the 3 pre-trained models using CNN-dailymail dataset.}	
\label{fig:output1} 
\end{figure*}
\begin{figure*}[tbp] 
\centering 
\begin{tabular}{ |p{12.2cm}|} 		\hline
\textbf{Original dialogue:} Shelly This year I'm volunteering at the food shelter! Tracy Good 4 u! Jody Gr8! Shelly How about u? Any volunteer work? Tracy Nah. Not into that. Jody Sure! Every year I do some charity 4 Xmas.\\\hline 	    
{\bf Human-written summary:} Shelly is voluntering at a food shelter and asks if others do some volunteer work. Tracy is not into that but Jody always does some charity for Christmas. \\\hline 	    
{\bf Predicted summary with google/pegasus-cnn-dailymail:} Tracy Good 4 u! Jody Gr8!  \\\hline 	    
{\bf Predicted summary with T5-base:} Tracy Nah. Not into that. I do some charity 4 Xmas work .\\\hline 	 
{\bf Predicted summary with facebook/bart-large-cnn:} Tracy and Jody are doing some charity work for Christmas. Jody is volunteering at a food shelter. Tracy is not into that. \\\hline
\end{tabular} 	
\caption{Original dialogue, human-written summary, and the summaries produced by the 3 pre-trained models using SAMSung dataset.}	
\label{fig:output2} 
\end{figure*}

\begin{figure*}[!tbp] 
\centering 
\begin{tabular}{ |p{12.2cm}|} \hline
\textbf{Bill text (truncated):} SECTION 1. SHORT TITLE. This Act may be cited as ``Small Business Tax Modernization Act of 2004''. SEC. 2. UNIFIED PASS THRU ENTITY REGIME. a Termination of S Corporation Status. 1 No new s corporation elections. Subsection a of section 1362 of the Internal Revenue Code of 1986 is amended by adding at the end the following new paragraph `` 3 Termination of authority to make election. No election may be made under paragraph 1 for any taxable year beginning after December 31 2004.''. 2 Termination of status. Subsection d of section 1362 of such Code relating to termination is amended by adding at the end the following new paragraph `` 4 Treatment as partnership after 2014. An election under subsection a `` A shall not be effective for any taxable year beginning after December 31 2014 and `` B shall be treated as an election under section 7701 a 2 B for taxable years beginning after such date.''. 3 Effective date. The amendments made  .....\\\hline 	    
{\bf Human-written summary:} Small Business Modernization Act of 2004 Amends the Internal Revenue Code to 1 terminate subchapter S corporation elections after 2004 and subchapter S status after 2014 and to allow privately held domestic corporations in lieu of electing subchapter S treatment to elect to be treated as partnerships for tax purposes 2 set forth rules for the tax treatment of former subchapter S corporations electing partnership status and 3 exclude from net earnings from self employment partnership income attributable to capital. \\\hline 	    
{\bf Predicted summary with google/pegasus-cnn-dailymail:} No election may be made under paragraph 1 for any taxable year beginning after December 31 2004 .<n>An eligible corporation may elect to be treated as a partnership for purposes of this title .<n>No gain or loss shall be recognized to the corporation or the shareholders by reason of an election under clause i and  II except in the case of an election made by a S corporation after the end of the recognition period.\\\hline 	    
{\bf Predicted summary with T5-base:} No election may be made under paragraph 1 for any taxable year beginning after December 31 2004''. 2 Termination of status. Subsection d of section 1362 of such Code relating to termination is amended by adding at the end the following new paragraph  4 Treatment as partnership after 2014 . b Election by certain corporations to be taxed as partnership .\\\hline 	   
{\bf Predicted summary with facebook/bart-large-cnn:} This Act may be cited as the ''Small Business Tax Modernization Act of 2004'' No new s corporation elections are allowed. An eligible corporation may elect to be treated as a partnership for purposes of this title.\\ \hline
\end{tabular} 	
\caption{Original bill text, human-written summary, and the summaries produced by the 3 pre-trained models using BillSum dataset.}	
\label{fig:output3} 
\end{figure*}

\vspace{-8em}

\vspace*{4cm}

\section{Conclusion}
 In this paper, we observed outputs from different pre-trained models using different datasets. While using CNN-dailymail dataset and SAMSum dataset,  google/pegasus-cnn-dailymail model and facebook/bart-large-cnn model provides better results than T5-base model as per the f-score of ROUGE-1, ROUGE-2, and ROUGE-L performance metrics. While using BillSum dataset google/pegasus-cnn-dailymail  model provides better results than T5-base model and facebook/bart-large-cnn model as per the f-measure of ROUGE-1, ROUGE-2, and ROUGE-L performance metrics. 
 In the future, we aim to fine-tune the pre-trained models on the given datasets and examine if the performance is improved. We also plan to apply these models to documents in scholarly domain as an extension to our work in \cite{rehman2021automatic}.
\bibliographystyle{IEEEtran}
\bibliography{ref}

\begin{thebibliography}{10}
\providecommand{\url}[1]{#1}
\csname url@samestyle\endcsname
\providecommand{\newblock}{\relax}
\providecommand{\bibinfo}[2]{#2}
\providecommand{\BIBentrySTDinterwordspacing}{\spaceskip=0pt\relax}
\providecommand{\BIBentryALTinterwordstretchfactor}{4}
\providecommand{\BIBentryALTinterwordspacing}{\spaceskip=\fontdimen2\font plus
\BIBentryALTinterwordstretchfactor\fontdimen3\font minus
  \fontdimen4\font\relax}
\providecommand{\BIBforeignlanguage}[2]{{%
\expandafter\ifx\csname l@#1\endcsname\relax
\typeout{** WARNING: IEEEtran.bst: No hyphenation pattern has been}%
\typeout{** loaded for the language `#1'. Using the pattern for}%
\typeout{** the default language instead.}%
\else
\language=\csname l@#1\endcsname
\fi
#2}}
\providecommand{\BIBdecl}{\relax}
\BIBdecl

\bibitem{luhn1958automatic}
H.~P. Luhn, ``The automatic creation of literature abstracts,'' \emph{IBM
  Journal of Research and Development}, vol.~2, no.~2, pp. 159--165, 1958.

\bibitem{radev2002introduction}
D.~R. Radev, E.~Hovy, and K.~McKeown, ``Introduction to the special issue on
  summarization,'' \emph{Computational linguistics}, vol.~28, no.~4, pp.
  399--408, 2002.

\bibitem{knight2002summarization}
K.~Knight and D.~Marcu, ``Summarization beyond sentence extraction: A
  probabilistic approach to sentence compression,'' \emph{Artificial
  Intelligence}, vol. 139, no.~1, pp. 91--107, 2002.

\bibitem{jing2000cut}
H.~Jing and K.~McKeown, ``Cut and paste based text summarization,'' in
  \emph{1st Meeting of the North American Chapter of the Association for
  Computational Linguistics}, 2000.

\bibitem{genest2012fully}
P.-E. Genest and G.~Lapalme, ``Fully abstractive approach to guided
  summarization,'' in \emph{Proceedings of the 50th Annual Meeting of the
  Association for Computational Linguistics (Volume 2: Short Papers)}, 2012,
  pp. 354--358.

\bibitem{hermann2015teaching}
K.~M. Hermann, T.~Kocisky, E.~Grefenstette, L.~Espeholt, W.~Kay, M.~Suleyman,
  and P.~Blunsom, ``Teaching machines to read and comprehend,'' \emph{Advances
  in neural information processing systems}, vol.~28, pp. 1693--1701, 2015.

\bibitem{nallapati2016abstractive}
R.~Nallapati, B.~Zhou, C.~Gulcehre, B.~Xiang \emph{et~al.}, ``Abstractive text
  summarization using sequence-to-sequence rnns and beyond,'' \emph{arXiv
  preprint arXiv:1602.06023}, 2016.

\bibitem{gliwa2019samsum}
B.~Gliwa, I.~Mochol, M.~Biesek, and A.~Wawer, ``Samsum corpus: A
  human-annotated dialogue dataset for abstractive summarization,'' \emph{arXiv
  preprint arXiv:1911.12237}, 2019.

\bibitem{kornilova2019billsum}
A.~Kornilova and V.~Eidelman, ``Billsum: A corpus for automatic summarization
  of us legislation,'' \emph{arXiv preprint arXiv:1910.00523}, 2019.

\bibitem{lin2004rouge}
C.-Y. Lin, ``{ROUGE:} a package for automatic evaluation of summaries,'' in
  \emph{Text Summarization Branches Out}, 2004, pp. 74--81.

\bibitem{papineni2002bleu}
K.~Papineni, S.~Roukos, T.~Ward, and W.-J. Zhu, ``Bleu: a method for automatic
  evaluation of machine translation,'' in \emph{Proceedings of the 40th annual
  meeting of the Association for Computational Linguistics}, 2002, pp.
  311--318.

\bibitem{bahdanau2014neural}
D.~Bahdanau, K.~Cho, and Y.~Bengio, ``Neural machine translation by jointly
  learning to align and translate,'' \emph{arXiv preprint arXiv:1409.0473},
  2014.

\bibitem{luong2015effective}
M.-T. Luong, H.~Pham, and C.~D. Manning, ``Effective approaches to
  attention-based neural machine translation,'' \emph{arXiv preprint
  arXiv:1508.04025}, 2015.

\bibitem{see2017get}
A.~See, P.~J. Liu, and C.~D. Manning, ``Get to the point: Summarization with
  pointer-generator networks,'' \emph{arXiv preprint arXiv:1704.04368}, 2017.

\bibitem{sutskever2014sequence}
I.~Sutskever, O.~Vinyals, and Q.~V. Le, ``Sequence to sequence learning with
  neural networks,'' in \emph{Advances in neural information processing
  systems}, 2014, pp. 3104--3112.

\bibitem{lin2018global}
J.~Lin, X.~Sun, S.~Ma, and Q.~Su, ``Global encoding for abstractive
  summarization,'' \emph{arXiv preprint arXiv:1805.03989}, 2018.

\bibitem{shi2021neural}
T.~Shi, Y.~Keneshloo, N.~Ramakrishnan, and C.~K. Reddy, ``Neural abstractive
  text summarization with sequence-to-sequence models,'' \emph{ACM Transactions
  on Data Science}, vol.~2, no.~1, pp. 1--37, 2021.

\bibitem{aksenov2020abstractive}
D.~Aksenov, J.~Moreno-Schneider, P.~Bourgonje, R.~Schwarzenberg, L.~Hennig, and
  G.~Rehm, ``Abstractive text summarization based on language model
  conditioning and locality modeling,'' \emph{arXiv preprint arXiv:2003.13027},
  2020.

\bibitem{zhang2020pegasus}
J.~Zhang, Y.~Zhao, M.~Saleh, and P.~Liu, ``Pegasus: Pre-training with extracted
  gap-sentences for abstractive summarization,'' in \emph{International
  Conference on Machine Learning}.\hskip 1em plus 0.5em minus 0.4em\relax PMLR,
  2020, pp. 11\,328--11\,339.

\bibitem{raffel2019exploring}
C.~Raffel, N.~Shazeer, A.~Roberts, K.~Lee, S.~Narang, M.~Matena, Y.~Zhou,
  W.~Li, and P.~J. Liu, ``Exploring the limits of transfer learning with a
  unified text-to-text transformer,'' \emph{arXiv preprint arXiv:1910.10683},
  2019.

\bibitem{lewis2019bart}
M.~Lewis, Y.~Liu, N.~Goyal, M.~Ghazvininejad, A.~Mohamed, O.~Levy, V.~Stoyanov,
  and L.~Zettlemoyer, ``Bart: Denoising sequence-to-sequence pre-training for
  natural language generation, translation, and comprehension,'' \emph{arXiv
  preprint arXiv:1910.13461}, 2019.

\bibitem{rehman2021automatic}
T.~Rehman, D.~K. Sanyal, S.~Chattopadhyay, P.~K. Bhowmick, and P.~P. Das,
  ``Automatic generation of research highlights from scientific,'' in \emph{2nd
  Workshop on Extraction and Evaluation of Knowledge Entities from Scientific
  Documents (EEKE'21), collocated with JCDL'21}, 2021.

\end{thebibliography}
\end{document}